%% file: paper.tex
\documentclass[11pt,a4paper]{article}
\usepackage[hyperref]{emnlp-ijcnlp-2019}
\usepackage{times}
\usepackage{latexsym}
\usepackage{hyperref}
\usepackage{enumerate}
\usepackage{graphicx}
\usepackage{caption}
\usepackage{booktabs}
\usepackage{array, makecell}
\usepackage{float}
\usepackage{pgfplots}
\usepackage{pgfplotstable}
\usetikzlibrary[patterns]
\usepackage{amsmath} 
\usepackage{siunitx}

\DeclareMathOperator*{\argmax}{arg\,max}

\aclfinalcopy 

\input{macro.tex}

\title{Simple and Effective Noisy Channel Modeling for Neural Machine Translation}

\author{Kyra Yee\fair{} \quad Nathan Ng\fair{} \quad Yann N. Dauphin\google{}$^\dagger$ \quad Michael Auli\fair{} \\
\fair{}Facebook AI Research \\
\google{}Google Brain
}

\date{}

\begin{document}
\maketitle

\renewcommand* {\thefootnote}{\fnsymbol{footnote}}
\footnotetext{$\dagger$ Work done while at Facebook AI Research.}
\renewcommand*{\thefootnote}{\arabic{footnote}}

\begin{abstract}
Previous work on neural noisy channel modeling relied on latent variable models that incrementally process the source and target sentence.
This makes decoding decisions based on partial source prefixes even though the full source is available.
We pursue an alternative approach based on standard sequence to sequence models which utilize the entire source. These models perform remarkably well as channel models, even though they have neither been trained on, nor designed to factor over incomplete target sentences.
Experiments with neural language models trained on billions of words show that noisy channel models can outperform a direct model by up to 3.2 BLEU on WMT'17 German-English translation.
We evaluate on four language-pairs and our channel models consistently outperform strong alternatives such right-to-left reranking models and ensembles of direct models.\footnote{We release code and pre-trained models at \href{https://github.com/pytorch/fairseq/tree/master/examples/noisychannel}{https://github.com/pytorch/fairseq}}
\end{abstract}

\section{Introduction}
 
Sequence to sequence models directly estimate the posterior probability of a target sequence $y$ given a source sequence $x$~\citep{sutskever2014sequence,bahdanau2015neural,gehring2017convs2s,vaswani2017transformer} and can be trained with pairs of source and target sequences.
Unpaired sequences can be leveraged by data augmentation schemes such as back-translation, but direct models cannot naturally take advantage of unpaired data~\citep{sennrich2016bt,edunov2018bt}.

The noisy channel approach is an alternative which is used in statistical machine translation~\citep{brown1993mathematics,koehn2003statistical}.
It entails a channel model probability $p(x|y)$ that operates in the reverse direction as well as a language model probability $p(y)$.
The language model can be estimated on unpaired data and can take a separate form to the channel model. 
Noisy channel modeling mitigates explaining away effects that result in the source being ignored for highly likely output prefixes~\citep{klein2001explain}.

Previous work on neural noisy channel modeling relied on a complex latent variable model that incrementally processes source and target prefixes~\citep{yu2017neuralnoisy}.
This trades efficiency for accuracy because their model performs significantly less well than a vanilla sequence to sequence model.
For languages with similar word order, it can be sufficient to predict the first target token based on a short source prefix, but for languages where word order differs significantly, we may need to take the entire source sentence into account to make a decision.

In this paper, we show that a standard sequence to sequence model is an effective parameterization of the channel probability.
We train the model on full sentences and apply it to score the source given an incomplete target sentence.
This bases decoding decisions on scoring the entire source sequence and it is very simple and effective (\textsection\ref{sec:approach}).
We analyze this approach for various target prefix sizes and find that it is most accurate for large target context sizes. 
Our simple noisy channel approach consistently outperforms strong baselines such as online ensembles and left-to-right re-ranking setups (\textsection\ref{sec:experiments}).

\section{Approach}
\label{sec:approach}

The noisy channel approach applies Bayes' rule to model $p(y|x) = p(x|y) p(y)/ p(x)$, that is, the channel model $p(x|y)$ operating from the target to the source and a language model $p(y)$. We do not model $p(x)$ since it is constant for all $y$.
We compute the channel model probabilities as follows:
$$
p(x|y) = \sum_{j}^{|x|} \log p(x_j | x_0, x_1, ...x_{j-1},y)
$$
We refer to $p(y|x)$ as the direct model.
A critical choice in our approach is to model $p(x|y)$ with a standard Transformer architecture~\citep{vaswani2017transformer} as opposed to a model which factors over target prefixes~\citep{yu2017neuralnoisy}.
This setup presents a clear train/test mismatch: 
we train $p(x|y)$ on complete sentence-pairs and perform inference with incomplete target prefixes of varying size $k$, i.e., $p(x|y_1,\dots,y_k)$. 
However, we find standard sequence to sequence models to be very robust to this mismatch (\textsection\ref{sec:experiments}).

\paragraph{Decoding.}
To generate $y$ given $x$ with the channel model, we wish to compute $\argmax_y \log p(x|y) + \log p(y)$.
However, na\"ive decoding in this way is computationally expensive because the channel model $p(x|y)$ is conditional on each candidate target prefix.
For the direct model, it is sufficient to perform a single forward pass over the network parameterizing $p(y|x)$ to obtain output word probabilities for the entire vocabulary.
However, the channel model requires separate forward passes for each vocabulary word.

\paragraph{Approximation.}
To mitigate this issue, we perform a two-step beam search where the direct model pre-prunes the vocabulary~\citep{yu2017neuralnoisy}.
For beam size $k_1$, and for each beam, we collect $k_2$ possible next word extensions according to the direct model. 
Next, we score the resulting $k_1 \times k_2$ partial candidates with the channel model and then prune this set to size $k_1$.
Other approaches to pre-pruning may be equally beneficial but we adopt this approach for simplicity.\footnote{
Vocabulary selection can prune the vocabulary to a few hundred types with no loss in accuracy~\citep{lhostis2016vocab}.}
A downside of online decoding with the channel model approach is the high computational overhead: we need to invoke the channel model $k_1 \times k_2$  times compared to just $k_1$ times for the direct model.

\paragraph{Complexity.}
The model of~\citet{yu2017neuralnoisy} factorizes over source and target prefixes. 
During decoding, their model alternates between incrementally reading the target prefix and scoring a source prefix, resulting in a runtime of $O(n+m)$, where $n$ and $m$ are the source and target lengths, respectively.  
In comparison, our approach repeatedly scores the entire source for each target prefix, resulting in $O(mn)$ runtime. 
Although our approach has greater time complexity, the practical difference of scoring the tokens of a single source sentence instead of just one token is likely to be negligible on modern GPUs since all source tokens can be scored in parallel. 
Inference is mostly slowed down by the autoregressive nature of decoding. 
Scoring the entire source enables capturing more dependencies between the source and target, since the beginning of the target must explain the entire source, not just the beginning. 
This is especially critical when the word order between the source and target language varies considerably, and likely accounts for the lower performance of the direct model of~\citet{yu2017neuralnoisy} in comparison to a standard seq2seq model.

\paragraph{Model combinaton.}
Since the direct model needs to be evaluated for pre-pruning, we also include these probabilities in making decoding decisions.
We use the following linear combination of the channel model, the language model and the direct model for decoding:
\begin{equation}
\frac{1}{t} \log p(y|x) + \frac{\lambda_1}{s} \Big( \log p(x|y) + \log p(y) \Big)
\label{eq:channel}
\end{equation}
where $t$ is the length of the target prefix $y$, $s$ is the source sentence length and $\lambda$ is a tunable weight. 
Initially, we used separate weights for $p(x|y)$ and $p(y)$ but we found that a single weight resulted in the same accuracy and was easier to tune.
Scaling by $t$ and $s$ makes the scores of the direct and channel model comparable to each other throughout decoding.
In n-best re-ranking, we have complete target sentences which are of roughly equal length and therefore do not use per word scores.\footnote{Reranking experiments are also based on separate tunable weights for the LM and the channel model. However, results are comparable to a single weight.}

\section{Experiments}
\label{sec:experiments}

\paragraph{Datasets.}
For English-German (En-De) we train on WMT'17 data, validate on news2016 and test on news2017.
For reranking, we train models with a 40K joint byte pair encoding vocabulary (BPE; \citealt{sennrich2016bpe}). 
To be able to use the language model during online decoding, we use the vocabulary of the langauge model on the target side. 
For the source vocabulary, we learn a 40K byte pair encoding on the source portion of the bitext; we find using LM and bitext vocabularies give similar accuracy.
For Chinese-English (Zh-En), we pre-process WMT'17 data following~\citet{hassan2018parity}, we develop on dev2017 and test on news2017. 
For IWSLT'14 De-En we follow the setup of~\citet{edunov2018classical} and measure case-sensitive tokenized BLEU.
For WMT De-En, En-De and Zh-En we measure detokenized BLEU~\citep{post:sacre:2018}.

\paragraph{Language Models.}
We train two big Transformer language models with 12 blocks~\citep{baevski2018adp}: one on the German newscrawl data distributed by WMT'18 comprising 260M sentences and another one on the English newscrawl data comprising 193M sentences.
Both use a BPE vocabulary of 32K types. 
We train on 32 Nvidia V100 GPUs with 16-bit floating point operations~\citep{ott:scaling:2018} and training took four days.

\paragraph{Sequence to Sequence Model training.}

For En-De, De-En, Zh-En we use big Transformers and for IWSLT De-En a base Transformer~\citep{vaswani2017transformer} as implemented in fairseq~\citep{ott2019fairseq}. 
For online decoding experiments, we do not share encoder and decoder embeddings since the source and target vocabularies were learned separately. 
We report average accuracy of three random initializations of a each configuration.
We generally use $k_1=5$ and $k_2=10$.
We tune $\lambda_1$, and a length penalty on the validation set.

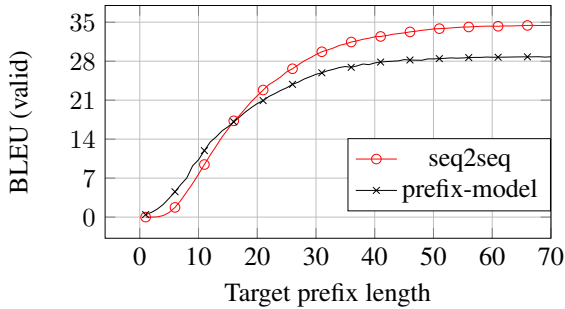
\begin{figure}[t]
\begin{center}
\resizebox{1\columnwidth}{!}{
\begin{tikzpicture}
\begin{axis}[
width=1.05\columnwidth,
height=0.65\columnwidth,
xmode=linear,
xlabel=Target prefix length,
ylabel=BLEU (valid),
xtick={0,10,20,30,40,50,60,70},
xticklabels={0,10,20,30,40,50,60,70},
ytick={0,7,14,21,28,35,42},
ymax=38,
xmax=70,
mark repeat=5,
legend style={at={(1.0,0.43)}},
grid=both]
\addplot[red,mark=o] table [y=seq2seq,x=len]{data/prefix_gen_wmt16_hyphen_splitting.dat};
\addplot[black,mark=x] table [y=prefix,x=len]{data/prefix_gen_wmt16_hyphen_splitting.dat};

\addlegendentry{seq2seq}
\addlegendentry{prefix-model}

\end{axis}
\end{tikzpicture}
}
\caption{Comparison of two channel models: a standard seq2seq model trained on full sentence-pairs and a model trained on all possible target prefixes with the full source (prefix-model). 
We measure accuracy of predicting the full source with increasing target prefixes for both models.
Results are on news2016.
\label{fig:genprefixes}}

\end{center}
\end{figure}

\subsection{Simple Channel Model}
 
We first motivate a standard sequence to sequence model to parameterize $p(x|y)$ as opposed to a model that is trained to operate over prefixes.
We train Transformer models to translate from the target to the source (En-De) and compare two variants: i) a standard sequence to sequence model trained to predict full source sentences based on full targets (seq2seq).
ii) a model trained to predict the full source based on a prefix of the target; we train on all possible prefixes of a target sentence, each paired with the full source  (prefix-model).

Figure~\ref{fig:genprefixes} shows that the prefix-model performs slightly better for short target prefixes but this advantage disappears after 15 tokens.
On full target sentences seq2seq outperforms the prefix model by 5.7 BLEU.
This is likely because the prefix-model needs to learn how to process both long and short prefixes which results in lower accuracy.
The lower performance on long prefixes is even more problematic considering our subsequent finding that channel models perform over-proportionally well on long target prefixes (\textsection\ref{sec:analysis}).
The seq2seq model has not been trained to process incomplete targets but empirically it provides a simple and effective parameterization of $p(x|y)$.

\subsection{Effect of Scoring the Entire Source Given Partial Target Prefixes}
\label{sec:prefix_source_rerrank}

The model of~\citep{yu2017neuralnoisy} uses a latent variable to incrementally score the source for prefixes of the target. Although this results in a faster run time, the model makes decoding decisions based on source prefixes even though the full source is available.
In order to quantify the benefit of scoring the entire source instead of a learned prefix length, we simulate different fractions of the source and target in an n-best list reranking setup.

\begin{figure}[t]
\begin{center}
\resizebox{1\columnwidth}{!}{
\begin{tikzpicture}
\begin{axis}[
legend style={font=\footnotesize},
width=1.05\columnwidth,
height=0.75\columnwidth,
xmode=linear,
xlabel=Target prefix length,
ylabel=BLEU score,
ymax=24.5,
legend pos= north west,
grid=both]
;
\addplot[red, mark=o] table [y=src0.2, x=prefix]{data/source_prefix_zh_en.dat};
\addplot[orange, mark=+] table [y=src0.4, x=prefix]{data/source_prefix_zh_en.dat};
\addplot[green, mark=square] table [y=src0.6, x=prefix]{data/source_prefix_zh_en.dat};
\addplot[blue, mark=x] table [y=src0.8, x=prefix]{data/source_prefix_zh_en.dat};
\addplot[purple, mark=*] table [y=src1, x=prefix]{data/source_prefix_zh_en.dat};

\addlegendentry{src 0.2}
\addlegendentry{src 0.4}
\addlegendentry{src 0.6}
\addlegendentry{src 0.8}
\addlegendentry{src 1}
\end{axis}
\end{tikzpicture}
}
\caption{For any given target prefix fraction, scoring the entire source has the best or comparable performance compared to other source prefixes. We show detokenized BLEU on the dev set of WMT17 Zh-En with beam 50.
\label{fig:source_prefix_zh_en}}
\end{center}
\end{figure}
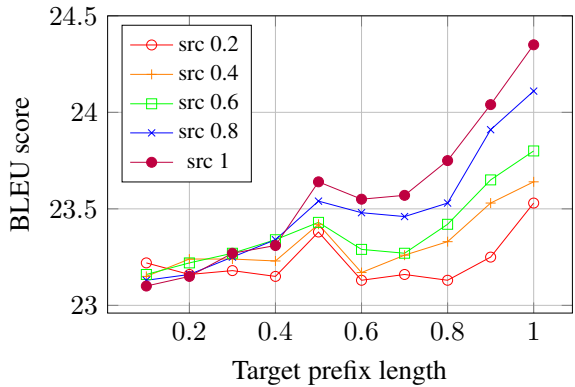

The n-best list is generated by the direct model and we re-rank the list in setups where we only have a fraction of the candidate hypothesis and the source sentence. 
We report BLEU of the selected full candidate hypothesis.
  
Figure~\ref{fig:source_prefix_zh_en} shows that for any given fraction of the target, scoring the entire source (src 1) has better or comparable performance than all other source prefix lengths. 
It is therefore beneficial to have a channel model that scores the entire source sentence.

\begin{table}[t]
\centering
\begin{tabular}{lrr}
\toprule
& news2016 & news2017 \\ \midrule
\baseline{} & 39.0 & 34.3  \\ 
\ens{} & 40.0 & 35.3 \\ 
\fwlm{} & 39.8 & 35.2 \\
\fwbwlm{} & 41.0 & 36.2 \\
- per word scores & 40.0 & 35.1 \\
\bottomrule
\end{tabular}
\caption{Online decoding accuracy for a direct model (\baseline{}), ensembling two direct models (\ens{}) and the channel approach (\fwbwlm{}). We ablate the impact of using per word scores. Results are on WMT De-En. 
Table~\ref{tab:onlinegenstdev} in the appendix shows standard deviations.}
\label{tab:onlinegen}
\end{table}

\subsection{Online Decoding}

Next, we evaluate online decoding with a noisy channel setup compared to just a direct model (\baseline{}) as well as an ensemble of two direct models (\ens{}).
Table~\ref{tab:onlinegen} shows that adding a language model to \baseline{} (\fwlm{}) gives a good improvement~\citep{gulcehre2015using} over a single direct model but ensembling two direct models is slightly more effective (\ens{}).
The noisy channel approach (\fwbwlm{}) improves by 1.9 BLEU over \baseline{} on news2017 and by 0.9 BLEU over the ensemble.
Without per word scores, accuracy drops because the direct model and the channel model are not balanced and their weight shifts throughout decoding. 
Our simple approach outperforms strong online ensembles which illustrates the advantage over incremental architectures~\citep{yu2017neuralnoisy} that do not match vanilla seq2seq models by themselves.

\subsection{Analysis}
\label{sec:analysis}

\begin{figure}[t]
\begin{center}
\resizebox{1\columnwidth}{!}{
\begin{tikzpicture}
\begin{axis}[
legend style={font=\footnotesize},
width=1.05\columnwidth,
height=0.6\columnwidth,
xmode=linear,
xlabel=Target prefix length,
ylabel=BLEU score,
xtick={0,10,20,30,40,50},
xticklabels={0,10,20,30,40,50},
ymax=42,
legend pos= north west,
grid=both]
\addplot[red, mark=+] table [y=channel, x=len]{data/prefixvaliddetok_hyphensplitting.dat};
\addplot[green, mark=o] table [y=fw_lm, x=len]{data/prefixvaliddetok_hyphensplitting.dat};
\addplot[purple, mark=x] table [y=fw_ens, x=len]{data/prefixvaliddetok_hyphensplitting.dat};

\addlegendentry{\fwbwlm{}}
\addlegendentry{\fwlm{}}
\addlegendentry{\ens{}}

\end{axis}
\end{tikzpicture}
}
\caption{Impact of target prefix length for the channel model (\fwbwlm{}), direct model + LM (\fwlm{}) and a direct ensemble (\ens{}). 
We show detokenized BLEU on WMT De-En news2016 with beam 10.
\label{fig:prefix}}
\end{center}
\end{figure}
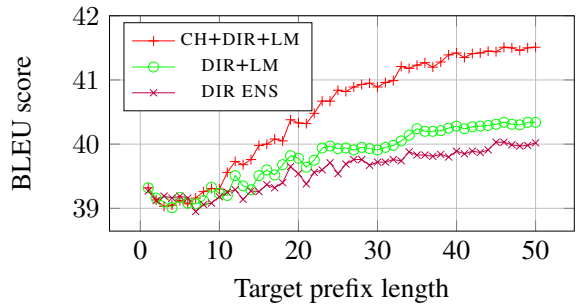

Using the channel model in online decoding enables searching a much larger space compared to n-best list re-ranking.
However, online decoding is also challenging because the channel model needs to score the entire source sequence given a partial target which can be hard.
To measure this, we simulate different target prefix lengths in an n-best list re-ranking setup. 
The n-best list is generated by the direct model and we re-rank it for different target prefixes of the candidate hypothesis.
As in \ref{sec:prefix_source_rerrank}, we measure BLEU of the selected full candidate hypothesis.
Figure~\ref{fig:prefix} shows that the channel model enjoys much larger benefits from more target context than re-ranking with just the direct model and an LM (\fwlm{}) or re-ranking with a direct ensemble (\ens{)}.
This experiment shows the importance of large context sizes for the channel approach to work well. 
It indicates that the channel approach may not be able to effectively exploit the large search space in online decoding due to the limited conditioning context provided by partial target prefixes.

\begin{table}[t]
\centering
\begin{tabular}{lrrrr}
\toprule
& 5 & 10 & 50 & 100  \\ \midrule
\baseline{} & 39.1 & 39.2 & 39.3  & 39.2  \\ 
\ens{} & 40.1 & 40.2 & 40.3  & 40.3    \\ 
\midrule
\fwlm{} & 40.0  & 40.2 &  40.6  & 40.7  \\
\lrrl{} & 39.7 & 40.1 & 40.8 & 40.8   \\
\lrrllm{} & 40.4 & 40.9 & 41.6 & 41.8  \\
\fwbw{} & 39.7 & 40.0 & 40.5 & 40.5 \\
\fwbwlm{} & 40.8 & 41.5 & 42.8  & 43.2  \\
\bottomrule
\end{tabular}
\caption{Re-ranking BLEU with different n-best list sizes on news2016 of WMT De-En. 
We compare to decoding with a direct model only (\baseline{}) and decoding with an ensemble of direct models (\ens{}). 
Table~\ref{tab:rerankbeamdev} in the appendix shows standard deviations.
}
\label{tab:rerankbeam}
\end{table}

\begin{table}[t]
\centering
\begin{tabular}{lrrrr}
\toprule
& \thead{WMT\\De-En} & \thead{WMT\\En-De} & \thead{WMT\\Zh-En} & \thead{IWSLT\\De-En} \\ \midrule
\baseline & 34.5 & 28.4 & 24.4  & 33.3 \\ 
\ens & 35.5 & 29.0 & 25.2  & 34.5 \\ 
\midrule
\fwlm{} & 36.0  & 29.4 &  24.9  & 34.2 \\
\lrrl{} & 35.7 & 29.3 & 25.3 & 34.4 \\
\lrrllm{} & 36.8 & 30.0 & 25.4 & 34.9 \\
\fwbw{} & 35.1 & 28.3 & 24.8 & 34.0\\
\fwbwlm{} & 37.7 & 30.5 & 25.6  & 35.5 \\
\bottomrule
\end{tabular}
\caption{Re-ranking accuracy with $k_1=50$ on four language directions on the respective test sets. See Table ~\ref{tab:reranktestdev} in the appendix for standard deviations.
}
\label{tab:reranktest}
\end{table}

\subsection{Re-ranking}

Next, we switch to n-best re-ranking where we have the full target sentence available compared to online decoding. 
Noisy channel model re-ranking has been used by the top ranked entries of the WMT 2019 news translation shared task for English-German, German-English, Englsh-Russian and Russian-English~\citep{ng2019fairwmt}.
We compare to various baselines including right-to-left sequence to sequence models which are a popular choice for re-ranking and regularly feature in successful WMT submissions~\citep{deng2018wmt,koehn2018wmt,junczysdowmunt2018wmt}.

Table~\ref{tab:rerankbeam} shows that the noisy channel model outperforms the baseline (\baseline{}) by up to 4.0 BLEU for very large beams, the ensemble by up to 2.9 BLEU (\ens{}) and the best right-to-left configuration by 1.4 BLEU (\lrrllm{}).
The channel approach improves more than other methods with larger n-best lists by adding 2.4 BLEU from $k_1=5$ to $k_1=100$. 
Other methods improve a lot less with larger beams, e.g., \lrrllm{} has the next largest improvement of 1.4 BLEU when increasing the beam size but this is still significantly lower than for the noisy channel approach. 
Adding a language model benefits all settings (\fwlm{}, \lrrllm{}, \fwbwlm{}) but the channel approach benefits most (\fwbw{} vs \fwbwlm{}).
The direct model with a language model (\fwlm{}) performs better than for online decoding, likely because the constrained re-ranking setup mitigates explaining away effects (cf. Table~\ref{tab:onlinegen}). 

Interestingly, both \fwbw{} or \fwlm{} give only modest improvements compared to \fwbwlm{}. 
Although previous work demonstrated that reranking with \fwbw{} can improve over \baseline{}, we show that the channel model is important to properly leverage the language model without suffering from explaining away effects~\citep{xu2018university,sogou_channel_reranking}. 
Test results on all language directions confirm that \fwbwlm{} performs best (Table~\ref{tab:reranktest}).

\section{Conclusion}

Previous work relied on incremental channel models which do not make use of the entire source even though it is available and, as we demonstrate, beneficial.
Standard sequence to sequence models are a simple parameterization for the channel probability that naturally exploits the entire source.
This parameterization outperforms strong baselines such as ensembles of direct models and right-to-left models.
Channel models are particularly effective with large context sizes and an interesting future direction is to iteratively refine the output while conditioning on previous contexts.

\bibliography{master}
\bibliographystyle{acl_natbib}

\appendix
\clearpage
\onecolumn

\section{Results with Standard Deviations}
\label{sec:appendix}

\begin{table}[h!]
\centering
\begin{tabular}{lrr}
\toprule
& news2016 & news2017 \\ \midrule
\baseline{} & $39.0 \pm $0.1 & $34.3 \pm $0.1  \\ 
\ens{} & $40.0 \pm $0.0 & $35.3 \pm $0.1 \\ 
\fwlm{} & $39.8 \pm $0.1 & $35.2 \pm $0.3 \\
\fwbwlm{} & $41.0 \pm $0.0 & $36.2 \pm $0.2 \\
- per word scores & $40.0 \pm $0.0 & $35.1 \pm $0.2 \\
\bottomrule
\end{tabular}
\caption{Online decoding accuracy for a direct model (\baseline{}), ensembling two direct models (\ens{}) and the channel approach (\fwbwlm{}). We ablate the impact of length normalization. Results are on news2017 of WMT De-En.}
\label{tab:onlinegenstdev}
\end{table}

\begin{table*}[h!]
\centering
\begin{tabular}{lrrrr}
\toprule
& 5 & 10 & 50 & 100  \\ \midrule
\baseline{} & $ 39.1 \pm 0.2 $ & $ 39.2 \pm 0.0 $ & $ 39.3 \pm 0.2  $ & $ 39.2 \pm 0.1  $ \\ 
\ens{} & $ 40.1  \pm 0.2 $ & $ 40.2 \pm 0.1 $ & $ 40.3 \pm 0.2  $ & $ 40.3  \pm 0.2 $  \\ 
\fwlm{} & $ 40.0 \pm 0.2  $ & $ 40.2 \pm 0.1 $ & $  40.6 \pm 0.2 $ & $ 40.7 \pm 0.1 $ \\
\lrrl{} & $ 39.7 \pm 0.1 $ & $ 40.1 \pm 0.2 $ & $ 40.8\pm 0.2 $ & $ 40.8 \pm 0.2 $  \\
\lrrllm{} & $ 40.4 \pm 0.2 $ & $ 40.9 \pm 0.2 $ & $ 41.6 \pm 0.2 $ & $ 41.8 \pm 0.2 $ \\
\fwbw{} & $ 39.7 \pm 0.2 $ & $ 40.0 \pm 0.2 $ & $ 40.5 \pm 0.0 $ & $ 40.5 \pm 0.1 $ \\
\fwbwlm{} & $ 40.8 \pm 0.2 $ & $ 41.52 \pm 0.1 $ & $ 42.8\pm 0.2  $ & $ 43.2 \pm 0.0 $ \\

\bottomrule
\end{tabular}
\caption{Re-ranking BLEU with different n-best list sizes on news2016 of WMT De-En. 
}
\label{tab:rerankbeamdev}
\end{table*}

\begin{table*}[h!]
\centering
\begin{tabular}{lrrrr}
\toprule
& \thead{WMT\\De-En} & \thead{WMT\\En-De} & \thead{WMT\\Zh-En} & \thead{IWSLT\\De-En} \\
\midrule
\baseline & $34.5 \pm  0.2$ & $28.4  \pm 0.1 $ & $24.4 \pm 0.1$  & $33.3 \pm 0.9 $\\ 
\ens & $ 35.5 \pm 0.1 $ & $ 29.0 \pm  0.1  $ & $ 25.2 \pm 0.2    $ & $ 34.5 \pm 0.3 $\\ 
\midrule
\fwlm{}   & $ 36.0 \pm 0.2  $ & $ 29.4 \pm 0.1  $ & $  24.9 \pm 0.3    $ & $ 34.2 \pm 0.8 $ \\
\lrrl{} & $ 35.7 \pm 0.3 $ & $ 29.3  \pm 0.0  $ & $ 25.3 \pm 0.3   $ & $ 34.4 \pm 0.6 $ \\
\lrrllm{} & $ 36.8 \pm 0.1  $ & $ 29.9 \pm 0.1  $ & $ 25.4 \pm 0.1   $ & $ 34.9 \pm 0.6   $ \\
\fwbw{} & $ 35.1 \pm 0.1  $ & $ 28.3 \pm 0.1  $ & $ 24.8 \pm 0.2  $ & $ 34.0 \pm 0.6 $ \\
\fwbwlm{} & $ 37.7  \pm 0.1  $ & $ 30.5 \pm 0.1  $ & $ 25.6 \pm 0.1  $ & $ 35.5 \pm 0.7 $ \\
\bottomrule
\end{tabular}
\caption{Re-ranking accuracy with $k_1=50$ on four language directions on the respective test sets. 
}
\label{tab:reranktestdev}
\end{table*}

\end{document}

%% file: macro.tex
\newcommand{\fair}{$^\bigtriangleup$}    
\newcommand{\google}{$^\bigtriangledown$}

\def\baseline{\textsc{dir}}
\def\ens{\textsc{dir ens}}

\def\fwlm{\textsc{dir+lm}}
\def\fwbw{\textsc{ch+dir}}

\def\fwbwlm{\textsc{ch+dir+lm}}

\def\lrrl{\textsc{dir+rl}}
\def\lrrllm{\textsc{dir+rl+lm}}

%% file: paper.bbl
\begin{thebibliography}{25}
\expandafter\ifx\csname natexlab\endcsname\relax\def\natexlab#1{#1}\fi

\bibitem[{Baevski and Auli(2018)}]{baevski2018adp}
Alexei Baevski and Michael Auli. 2018.
\newblock Adaptive input representations for neural language modeling.
\newblock \emph{arXiv}, abs/1809.10853.

\bibitem[{Bahdanau et~al.(2015)Bahdanau, Cho, and Bengio}]{bahdanau2015neural}
Dzmitry Bahdanau, Kyunghyun Cho, and Yoshua Bengio. 2015.
\newblock Neural machine translation by jointly learning to align and
  translate.
\newblock In \emph{Proc. of ICLR}.

\bibitem[{Brown et~al.(1993)Brown, Pietra, Pietra, and
  Mercer}]{brown1993mathematics}
Peter~F Brown, Vincent J~Della Pietra, Stephen A~Della Pietra, and Robert~L
  Mercer. 1993.
\newblock The mathematics of statistical machine translation: Parameter
  estimation.
\newblock \emph{Computational linguistics}, 19(2):263--311.

\bibitem[{Deng et~al.(2018)Deng, Cheng, Lu, Song, Wang, Wu, Yao, Zhang, Zhang,
  Zhang, Zhu, and Chen}]{deng2018wmt}
Yongchao Deng, Shanbo Cheng, Jun Lu, Kai Song, Jingang Wang, Shenglan Wu, Liang
  Yao, Guchun Zhang, Haibo Zhang, Pei Zhang, Changfeng Zhu, and Boxing Chen.
  2018.
\newblock Alibaba's neural machine translation systems for wmt18.
\newblock In \emph{Proc. of WMT}.

\bibitem[{Edunov et~al.(2018{\natexlab{a}})Edunov, Ott, Auli, and
  Grangier}]{edunov2018bt}
Sergey Edunov, Myle Ott, Michael Auli, and David Grangier. 2018{\natexlab{a}}.
\newblock Understanding back-translation at scale.
\newblock In \emph{Proc. of EMNLP}.

\bibitem[{Edunov et~al.(2018{\natexlab{b}})Edunov, Ott, Auli, Grangier, and
  Ranzato}]{edunov2018classical}
Sergey Edunov, Myle Ott, Michael Auli, David Grangier, and Marc'Aurelio
  Ranzato. 2018{\natexlab{b}}.
\newblock Classical structured prediction losses for sequence to sequence
  learning.
\newblock In \emph{Proc. of NAACL}.

\bibitem[{Gehring et~al.(2017)Gehring, Auli, Grangier, Yarats, and
  Dauphin}]{gehring2017convs2s}
Jonas Gehring, Michael Auli, David Grangier, Denis Yarats, and Yann~N Dauphin.
  2017.
\newblock {Convolutional Sequence to Sequence Learning}.
\newblock In \emph{Proc. of ICML}.

\bibitem[{Gulcehre et~al.(2015)Gulcehre, Firat, Xu, Cho, Barrault, Lin,
  Bougares, Schwenk, and Bengio}]{gulcehre2015using}
Caglar Gulcehre, Orhan Firat, Kelvin Xu, Kyunghyun Cho, Loic Barrault, Huei-Chi
  Lin, Fethi Bougares, Holger Schwenk, and Yoshua Bengio. 2015.
\newblock On using monolingual corpora in neural machine translation.
\newblock \emph{arXiv}, abs/1503.03535.

\bibitem[{Hassan et~al.(2018)Hassan, Aue, Chen, Chowdhary, Clark, Federmann,
  Huang, Junczys-Dowmunt, Lewis, Li, Liu, Liu, Luo, Menezes, Qin, Seide, Tan,
  Tian, Wu, Wu, Xia, Zhang, Zhang, and Zhou}]{hassan2018parity}
Hany Hassan, Anthony Aue, Chang Chen, Vishal Chowdhary, Jonathan Clark,
  Christian Federmann, Xuedong Huang, Marcin Junczys-Dowmunt, Will Lewis,
  Mu~Li, Shujie Liu, Tie-Yan Liu, Renqian Luo, Arul Menezes, Tao Qin, Frank
  Seide, Xu~Tan, Fei Tian, Lijun Wu, Shuangzhi Wu, Yingce Xia, Dongdong Zhang,
  Zhirui Zhang, and Ming Zhou. 2018.
\newblock Achieving human parity on automatic chinese to english news
  translation.
\newblock \emph{arXiv}, abs/1803.05567.

\bibitem[{Junczys-Dowmunt(2018)}]{junczysdowmunt2018wmt}
Marcin Junczys-Dowmunt. 2018.
\newblock Microsoft's submission to the wmt2018 news translation task: How i
  learned to stop worrying and love the data.
\newblock In \emph{Proc. of WMT}.

\bibitem[{Klein and Manning(2001)}]{klein2001explain}
Dan Klein and Christopher Manning. 2001.
\newblock Conditional structure versus conditional estimation in nlp.
\newblock In \emph{Proc. of EMNLP}.

\bibitem[{Koehn et~al.(2018)Koehn, Duh, and Thompson}]{koehn2018wmt}
Philipp Koehn, Kevin Duh, and Brian Thompson. 2018.
\newblock The jhu machine translation systems for wmt 2018.
\newblock In \emph{Proc. of WMT}.

\bibitem[{Koehn et~al.(2003)Koehn, Och, and Marcu}]{koehn2003statistical}
Philipp Koehn, Franz~Josef Och, and Daniel Marcu. 2003.
\newblock Statistical phrase-based translation.
\newblock In \emph{Proc. of NAACL}.

\bibitem[{L'Hostis et~al.(2016)L'Hostis, Grangier, and Auli}]{lhostis2016vocab}
Gurvan L'Hostis, David Grangier, and Michael Auli. 2016.
\newblock Vocabulary selection strategies for neural machine translation.
\newblock \emph{arXiv}, abs/1610.00072.

\bibitem[{Ng et~al.(2019)Ng, Yee, Baevski, Ott, Auli, and
  Edunov}]{ng2019fairwmt}
Nathan Ng, Kyra Yee, Alexei Baevski, Myle Ott, Michael Auli, and Sergey Edunov.
  2019.
\newblock Facebook fair's wmt19 news translation task submission.
\newblock In \emph{Proc. of WMT}.

\bibitem[{Ott et~al.(2019)Ott, Edunov, Baevski, Fan, Gross, Ng, Grangier, and
  Auli}]{ott2019fairseq}
Myle Ott, Sergey Edunov, Alexei Baevski, Angela Fan, Sam Gross, Nathan Ng,
  David Grangier, and Michael Auli. 2019.
\newblock fairseq: A fast, extensible toolkit for sequence modeling.
\newblock In \emph{Proc. of NAACL System Demonstrations}.

\bibitem[{Ott et~al.(2018)Ott, Edunov, Grangier, and Auli}]{ott:scaling:2018}
Myle Ott, Sergey Edunov, David Grangier, and Michael Auli. 2018.
\newblock Scaling neural machine translation.
\newblock In \emph{Proc. of WMT}.

\bibitem[{Post(2018)}]{post:sacre:2018}
Matt Post. 2018.
\newblock A call for clarity in reporting bleu scores.
\newblock \emph{arXiv}, abs/1804.08771.

\bibitem[{Sennrich et~al.(2016{\natexlab{a}})Sennrich, Haddow, and
  Birch}]{sennrich2016bt}
Rico Sennrich, Barry Haddow, and Alexandra Birch. 2016{\natexlab{a}}.
\newblock Improving neural machine translation models with monolingual data.
\newblock In \emph{Proc. of ACL}.

\bibitem[{Sennrich et~al.(2016{\natexlab{b}})Sennrich, Haddow, and
  Birch}]{sennrich2016bpe}
Rico Sennrich, Barry Haddow, and Alexandra Birch. 2016{\natexlab{b}}.
\newblock Neural machine translation of rare words with subword units.
\newblock In \emph{Proc. of ACL}.

\bibitem[{Sutskever et~al.(2014)Sutskever, Vinyals, and
  Le}]{sutskever2014sequence}
Ilya Sutskever, Oriol Vinyals, and Quoc~V Le. 2014.
\newblock {Sequence to Sequence Learning with Neural Networks}.
\newblock In \emph{Proc. of NIPS}.

\bibitem[{Vaswani et~al.(2017)Vaswani, Shazeer, Parmar, Uszkoreit, Jones,
  Gomez, Kaiser, and Polosukhin}]{vaswani2017transformer}
Ashish Vaswani, Noam Shazeer, Niki Parmar, Jakob Uszkoreit, Llion Jones,
  Aidan~N. Gomez, Lukasz Kaiser, and Illia Polosukhin. 2017.
\newblock {Attention Is All You Need}.
\newblock In \emph{Proc. of NIPS}.

\bibitem[{Wang et~al.(2017)Wang, Cheng, Jiang, Yang, Chen, Li, Shi, Wang, and
  Yang}]{sogou_channel_reranking}
Yuguang Wang, Shanbo Cheng, Liyang Jiang, Jiajun Yang, Wei Chen, Muze Li, Lin
  Shi, Yanfeng Wang, and Hongtao Yang. 2017.
\newblock Sogou neural machine translation systems for wmt17.
\newblock In \emph{Proceedings of the Second Conference on Machine
  Translation}, pages 410--415.

\bibitem[{Xu and Carpuat(2018)}]{xu2018university}
Weijia Xu and Marine Carpuat. 2018.
\newblock The university of maryland's chinese-english neural machine
  translation systems at wmt18.
\newblock In \emph{Proceedings of the Third Conference on Machine Translation:
  Shared Task Papers}, pages 535--540.

\bibitem[{Yu et~al.(2017)Yu, Blunsom, Dyer, Grefenstette, and
  Kocisk{\'{y}}}]{yu2017neuralnoisy}
Lei Yu, Phil Blunsom, Chris Dyer, Edward Grefenstette, and Tom{\'{a}}s
  Kocisk{\'{y}}. 2017.
\newblock The neural noisy channel.
\newblock In \emph{Proc. of ICLR}.

\end{thebibliography}
